\pdfoutput=1

\documentclass[11pt]{article}

\usepackage[final]{acl}

\usepackage{times}
\usepackage{latexsym}
\usepackage{amsmath}
\usepackage{mathtools}
\usepackage{colortbl}
\usepackage{float} 

\usepackage[T1]{fontenc}

\usepackage[utf8]{inputenc}

\usepackage{microtype}

\usepackage{inconsolata}

\usepackage{graphicx}
\usepackage{multirow}
\usepackage{booktabs}
\usepackage{amssymb}
\usepackage{pifont} 
\usepackage{xcolor} 
\usepackage{tcolorbox}
\usepackage[normalem]{ulem}
\useunder{\uline}{\ul}{}
\usepackage{lipsum}
\usepackage{geometry}
\usepackage{multicol}
\usepackage{titlesec}
\usepackage{caption}
\usepackage{makecell}
\usepackage{enumitem}

\title{Assessing LLM Reasoning Steps via Principal Knowledge Grounding}

\author{Hyeon Hwang$^1$ \hspace*{0.1cm}
    Yewon Cho$^1$ \hspace*{0.1cm}
    Chanwoong Yoon$^1$ \hspace*{0.1cm}
    Yein Park$^{1}$ \hspace*{0.1cm}
    Minju Song$^{1}$ \hspace*{0.1cm} \\
    \textbf{Kyungjae Lee}$^{2}$ \hspace*{0.1cm}
    \textbf{Gangwoo Kim}$^{3}$\thanks{Work done before joining the affiliation} \hspace*{0.1cm}
    \textbf{Jaewoo Kang}$^{1,4}$ \\
    Korea University$^1$ \hspace*{0.1cm} University of Seoul$^2$ \hspace*{0.1cm} AWS AI Labs$^3$ \hspace*{0.1cm} AIGEN Sciences$^4$ \\
    \{hyeon-hwang, yewon330,  kangj\}@korea.ac.kr
}

\begin{document}
\maketitle

\begin{abstract}
Step-by-step reasoning has become a standard approach for large language models (LLMs) to tackle complex tasks. 
While this paradigm has proven effective, it raises a fundamental question: \textit{How can we verify that an LLM's reasoning is accurately grounded in knowledge?}
To address this question, we introduce a novel evaluation suite that systematically assesses the knowledge grounding of intermediate reasoning. 
Our framework comprises three key components.
(1) \textsc{Principal Knowledge Collection}, a large-scale repository of atomic knowledge essential for reasoning. 
Based on the collection, we propose (2) knowledge-grounded evaluation metrics designed to measure how well models recall and apply prerequisite knowledge in reasoning. 
These metrics are computed by our (3) evaluator LLM, a lightweight model optimized for cost-effective and reliable metric computation.
Our evaluation suite demonstrates remarkable effectiveness in identifying missing or misapplied knowledge elements, providing crucial insights for uncovering fundamental reasoning deficiencies in LLMs.
Beyond evaluation, we demonstrate how these metrics can be integrated into preference optimization, showcasing further applications of knowledge-grounded evaluation. Our evaluation suite is publicly available.\footnote{ \href{https://github.com/dmis-lab/PKCollection}{https://github.com/dmis-lab/PKCollection}.}

\end{abstract}
\begin{figure}[t!]
\centering
\includegraphics[width=\columnwidth]{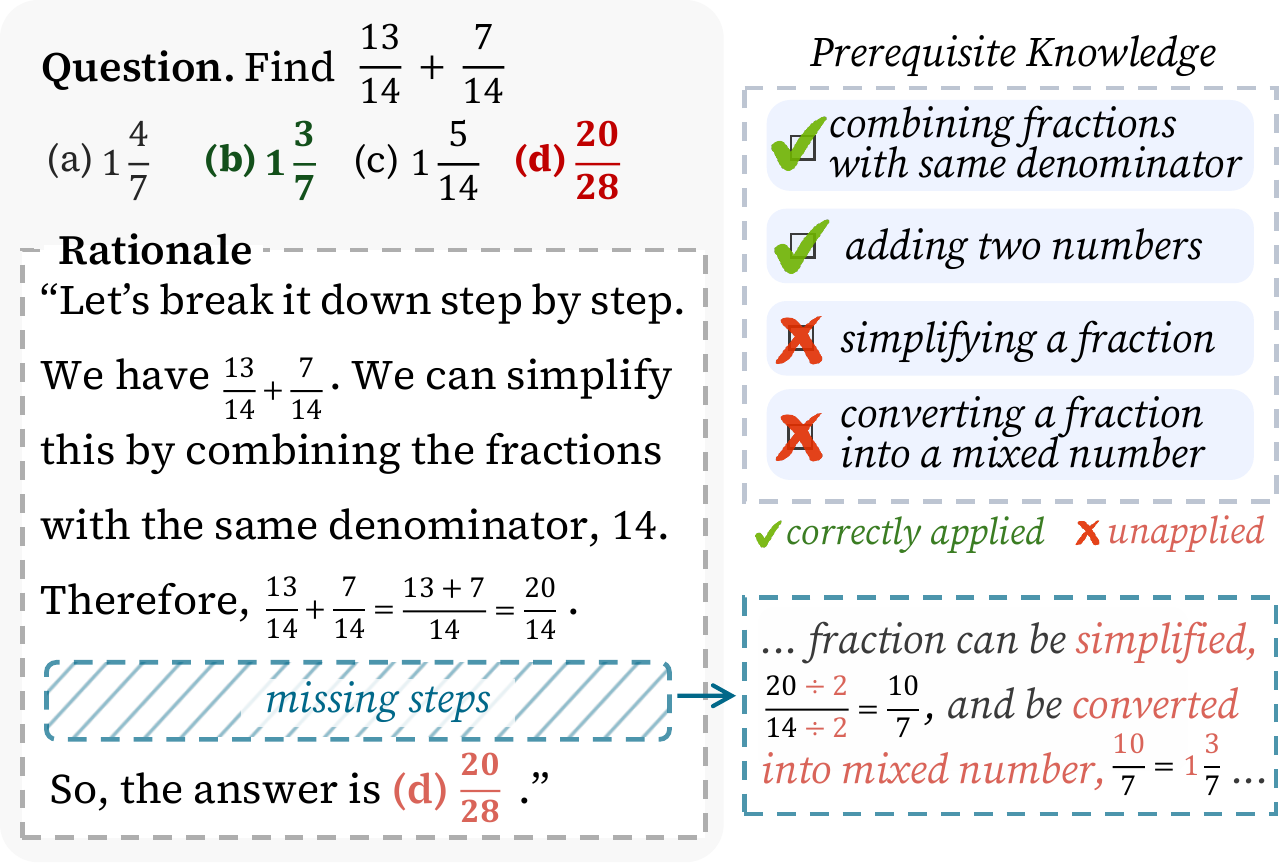}

\caption{
Example of a reasoning failure.
The model correctly applies the knowledge of ``\textit{combining fractions}'' but omits prerequisite steps such as ``\textit{simplifying a fraction},'' which leads to an incorrect answer.
This underscores the importance of evaluating reasoning steps through knowledge-grounded assessments.
}
\label{fig:motivation_example}
\vspace{-0.3cm}
\end{figure}
\section{Introduction}
Recently, large language models (LLMs)~\cite{openai2024reasoning, deepseekai2025deepseekr1incentivizingreasoningcapability} have demonstrated remarkable reasoning capabilities.
These models have achieved substantial performance gains by generating their own intermediate reasoning steps before arriving at a final answer.
However, such gains are often assessed solely through end-task metrics, which evaluates only the final output.
While this simplicity offers practical efficiency in evaluation, it provides limited insights into how reliably or meaningfully the model performs reasoning, failing to detect errors in intermediate steps.

To address this, prior works~\citep{prasad-etal-2023-receval, golovneva2023roscoe, chen-etal-2023-rev, hao2024llm} have focused on evaluating the validity of the intermediate reasoning steps that lead to the final output.
They evaluate the reasoning steps by identifying errors like hallucinations~\cite{lei2025hallucination} and logical flaws. Yet, these approaches do not adequately measure how a model utilizes the essential knowledge required for problem-solving, making it difficult to provide interpretable feedback on model behavior.
Figure~\ref{fig:motivation_example} exemplifies this limitation, showing how a model can apply partial knowledge correctly while overlooking other prerequisite steps.
While it correctly applies combining fractions, it neglects necessary knowledge, such as simplifying a fraction and converting the result into a mixed number.



To systematically assess how precisely a model utilizes essential  knowledge, we propose an evaluation suite consisting of the following three components:
We first construct (1) \textsc{Principal Knowledge Collection} (\textsc{PK Collection}), a large-scale resource of atomic knowledge essential for solving the target tasks. 
This knowledge is collected from multiple top-performing LLMs, clustered into coherent chunks, and eventually refined into principal knowledge for each chunk. 
In particular, we built this collection based on the MMLU benchmark~\cite{hendrycks2020measuring}, resulting in 112k principal knowledge units  (PK units).
Grounded in the \textsc{PK Collection}, we introduce (2) novel evaluation metrics, including knowledge-grounded precision, recall, and F1 score. 
The LLM-based evaluation process involves identifying the utilization of PK units within the predicted rationale. 
These metrics measure how precisely the model applies the knowledge and whether it recalls the necessary information accurately.

Furthermore, using proprietary LLMs for reasoning evaluation is costly and impractical at scale.
To overcome this, we develop (3) a light-weight LLM evaluator distilled from a  proprietary teacher LLM, achieving high agreement with reduced computational overhead.
We also validate \textsc{PK Collection} for factuality and relevance, confirming that it maintains high quality in both aspects.

To additionally explore the potential applications of our evaluation suite, we investigate its use in controlling reasoning conciseness with recent preference optimization techniques. 
Following the approach of \citet{pang2025iterative}, we adopt a framework that applies Direct Preference Optimization (DPO)~\cite{rafailov2023direct}. 
By selecting preferred samples based on our evaluation metrics, we observe that models could efficiently generate reasoning steps while maintaining end performance.
This result demonstrates that aligning LLM with our evaluation metrics effectively guides it to concisely reach the final result for solving the task.


Our experimental results and analysis demonstrate that our knowledge-grounded metrics effectively evaluate knowledge utilization and provide interpretable feedback on model behavior.
Moreover, reasoning preference optimization guided by our knowledge-grounded metric not only boosts performance but also enables control over the model's knowledge usage and token consumption.

Our contributions are summarized in threefold:

\begin{itemize}
    \item We construct \textsc{Principal Knowledge Collection}, a large-scale resource containing 112K principal knowledge for solving questions from diverse domains with high accuracy and relevance.
    \item We propose a novel evaluation system equipped with knowledge-grounded metrics and a distilled evaluator LLM, which assesses the utilization of principal knowledge in LLM reasoning.
    \item We extend our evaluation suite to enable more controllable model reasoning via preference optimization. Our analysis showcases how this approach encourages more concise reasoning, which also leads to reduced token consumption.
\end{itemize}
\section{Related Works}
\begin{figure*}[t]
\includegraphics[width=\textwidth]{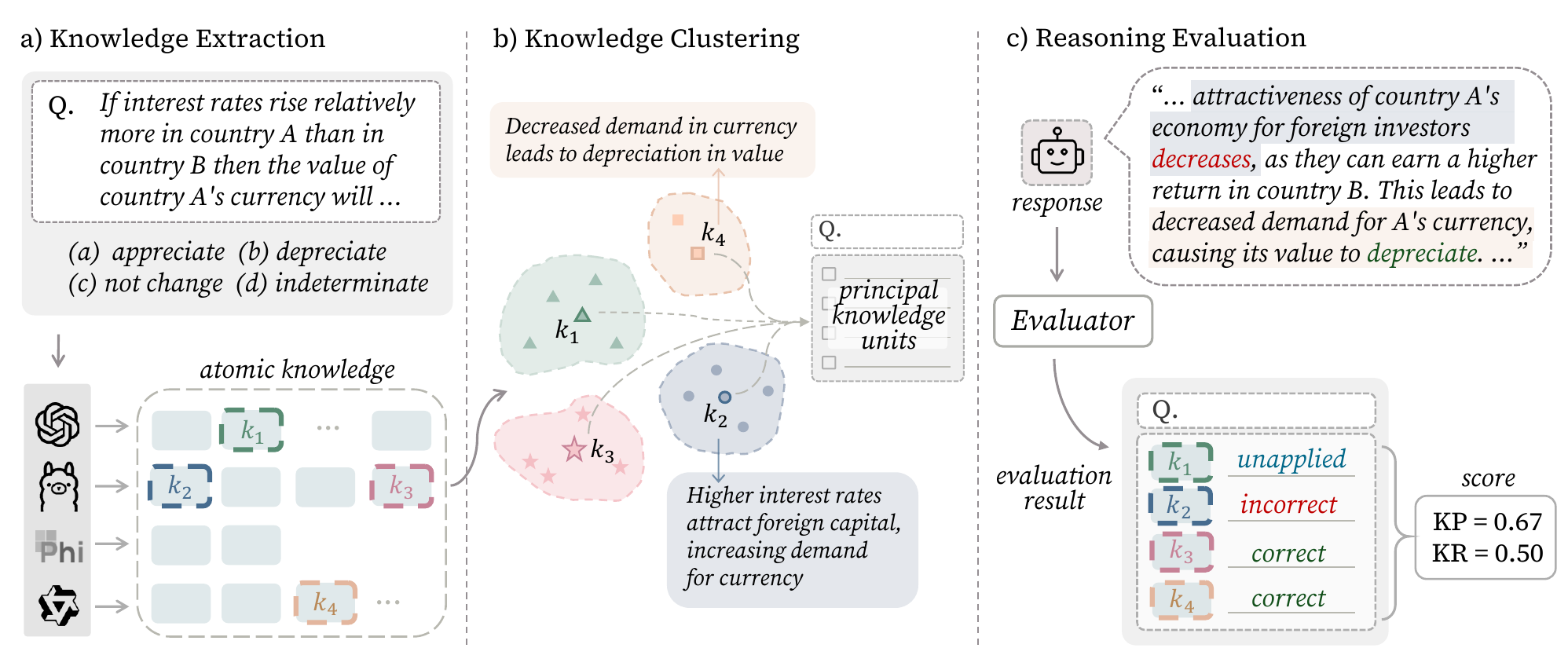}
\caption{
Overview of our evaluation suite for assessing LLM reasoning steps via knowledge grounding.
We first construct the \textsc{Principal Knowledge Collection} (\S~\ref{sec:kc_construction}) through a two-step process: 
Given a task, we (a) collect atomic knowledge crucial for task resolution from multiple top-performing LLMs. 
Subsequently, we (b) cluster these units into semantically coherent groups to obtain the principal knowledge for each cluster. 
Grounded on this collection, we (c) evaluate the model’s reasoning steps (\S~\ref{sec:eval_reasoning}) by measuring whether the model accurately retrieves and applies the principal knowledge using our proposed knowledge recall and precision metrics.
}
\label{fig:main_fig}
\vspace{-5mm}
\end{figure*}

\quad \textbf{LLM Reasoning Steps} Chain-of-Thought~\citep{wei2022chain} encourages the generation of intermediate reasoning steps, improving LLMs' ability to solve complex tasks. 
Self-consistency~\citep{wang2023selfconsistency} improves reliability by sampling multiple reasoning paths and selecting the most frequent answer through majority voting. 
Tree-of-Thought~\citep{yao2023tree} employs heuristic-based search strategies, such as depth-first and breadth-first search, to explore more effective reasoning paths.
Chain-of-Knowledge~\citep{li2024chainofknowledge} focuses on improving factuality by dynamically grounding reasoning steps with external sources.

\textbf{LLM-based Evaluation} Recent studies explore the use of LLMs as evaluators, automating response assessment, and reducing the dependence on human annotations ~\citep{zheng2023judging,DBLP:journals/corr/abs-2302-04166, lyu-etal-2024-knowtuning}, and also show strong correlation with human judgements~\citep{liusie-etal-2024-llm}. 
To mitigate dependence on proprietary language models, subsequent research has explored training models specifically designed for evaluation tasks~\citep{li2023generative, kim2024prometheus, kim-etal-2024-prometheus}.

\textbf{LLM Reasoning Evaluation} Recent advancements have proposed fine-grained methods for evaluating intermediate reasoning in LLMs. Beyond assessing sentence-level correctness, prior work has examined the alignment of intermediate steps with the final answer \citep{prasad-etal-2023-receval}, the relevance and hallucination rate as well as overall fluency \citep{golovneva2023roscoe}, and the utility of conditional information \citep{chen-etal-2023-rev}. However, despite these advances, there remains a lack of research on whether LLMs' reasoning steps are truly grounded in accurate and relevant knowledge, a key to interpretable and reliable model behavior.

\section{Evaluation Suite for Reasoning Steps Grounded on Principal Knowledge}
\label{sec:evaluation_suite}
To assess the reasoning steps via knowledge grounding, we propose a novel evaluation suite:
We first construct \textsc{Principal Knowledge Collection} (\S~\ref{sec:kc_construction}), a large-scale resource of atomic knowledge essential for solving the target tasks.
We also introduce novel evaluation metrics (\S~\ref{sec:eval_reasoning}) to measure the utilization of principal knowledge within the thinking process. 
To facilitate efficient evaluation, we further develop 
 an open-weight LLM evaluator (\S~\ref{sec:training_evaluator}) by distilling a proprietary LLM. Figure~\ref{fig:main_fig} shows the overview of our evaluation suite construction and evaluation.

\subsection{\textsc{Principal Knowledge Collection}}
\label{sec:kc_construction}
We construct \textsc{Principal Knowledge Collection}, which serves as the reference foundation for our evaluation suite.
We first collect atomic knowledge required for solving the target tasks (\S~\ref{subsec:collecting_units}). 
Subsequently, we segment them into coherent clusters, resulting in principal knowledge for each cluster (\S~\ref{subsec:refine_chunks}). 

\subsubsection{Collecting Atomic Knowledge}
\label{subsec:collecting_units}
To systematically identify and enumerate the principal knowledge for resolving the task, we collect atomic knowledge from top-performing LLMs. The generator LLMs we employed—GPT-4o (88.7\%), Llama3-70B-instruct (86.0\%), Qwen2.5-72B (86.1\%), and Phi-4 (84.8\%)—closely approach human expert performance (89.8\%) in the MMLU benchmark, effectively generating comprehensive and precise atomic knowledge even in complex domains.
Specifically, we prompt the LLM to generate a list of atomic knowledge that contributes to deriving the correct answer.\footnote{See detailed prompts in Table \ref{table:prompt_knowledge}}
Each atomic knowledge is generated and parsed as a self-contained statement that clearly expresses a relevant principle or concept. For example, given a task, ``\textit{When using the ideal gas law, what are the standard conditions for temperature and pressure?}'', a statement \textit{"Standard temperature is defined as \( 0^{\circ} \text{C} \) ( \( 273.15\, \text{K} \) )."} is generated.
We repeat this process, yielding a structured set of PK units for each instance: $\mathcal{K}=\{ k_{1}, k_{2}, \cdots, k_{|\mathcal{K}|}\}$.
We do not restrict the output number, 
allowing the model to identify and enumerate all relevant pieces of knowledge.

\subsubsection{Refining Knowledge Chunks}
\label{subsec:refine_chunks}
The collected chunks often contain duplicate or semantically equivalent units. 
To group them, we apply K-means clustering~\citep{macqueen1967some}, segmenting the PK units into distinct sub-groups based on their semantic similarity.
Specifically, we embed each PK unit using a transformer encoder,\footnote{\url{https://huggingface.co/nvidia/NV-Embed-v2}}~\citep{lee2025nvembed} and partition them into clusters, denoted as: $\mathcal{C} = \{\mathcal{C}_1, \mathcal{C}_2, \cdots, \mathcal{C}_{N}\}$ 
where the number of clusters $N$ is determined as the average number of PK units with an additive offset of 2.
To ensure a representative and concise knowledge set, we select a PK unit $k^*_j$ for each cluster $\mathcal{C}_j$ by identifying the one closest to the cluster centroid in the embedding space.
This eventually yields a refined knowledge collection consisting of unique and representative knowledge units, resulting in the \textsc{Principal knowledge collection} $\mathcal{K}^* = \{ k_1^*, k_2^*, \cdots, k_{|C|}^* \}$. Appendix~\ref{appendix:data_stats} shows the statistics of the resulting collection, and Appendix~\ref{appendix:pk_collection_example} provides an illustrative example of \textsc{Pk Collection} from MMLU college mathematics questions. Through this rigorous procedure, \textsc{PK Collection} effectively captures the core information of each cluster while preserving diversity, enabling efficient evaluation in subsequent steps.

\subsection{Evaluating Reasoning Steps}
\label{sec:eval_reasoning}
We assess the reasoning process of the prediction model based on our collection. 
Following the \textit{LLM-as-a-judge} paradigm ~\citep{zheng2023judging}, we leverage an LLM evaluator to assess the utilization of principal units.
This systematically evaluates whether the model correctly applies the necessary information with the following evaluation metrics.

\paragraph{Proposed Evaluation Metrics}
To measure the comprehensive utilization principal knowledge, we introduce novel evaluation metrics: 
To assess the correctness of the predicted rationale \(r\), we introduce (i) \textbf{Knowledge Precision}, which quantifies the proportion of correct PK units extracted from the rationale as follows:
$$
\mathrm{KP}(r) = \frac{1}{|\hat{\mathcal{K}}|} \sum_{i=1}^{|\hat{\mathcal{K}}|} \mathbb{I}(\text{Eval}(\hat{k}_i, \mathcal{K}^*))
$$
where the indicator function  $\mathbb{I}(\cdot)$  returns 1 if the condition is true, and 0 otherwise, and $\text{Eval}(\cdot, \mathcal{K}^*)$ denotes the evaluator LLM for assessing the correctness of each PK unit based on our collection $\mathcal{K}^*$.
This also serves as an indicator of hallucinated or unverified knowledge, capturing the extent to which the model introduces incorrect information. 
However, it is inherently limited to assessing only the knowledge explicitly present in the rationale and does not account for missing PK units required for a complete reasoning process.

To assess how many principal units appear in its reasoning, we define (ii) \textbf{Knowledge Recall}, which measures the proportion of PK units from our reference collection that appear in the predicted rationale. It is computed as follows:
$$
\mathrm{KR}(r) = \frac{1}{|\mathcal{K}^*|} \sum_{i=1}^{|\hat{\mathcal{K}}|} \mathbb{I}(\text{Eval}(\hat{k}_i, \mathcal{K}^*))
$$
 where the indicator function $\mathbb{I}(\cdot)$  returns 1 if the predicted PK unit  $\hat{k}_i$ is included in the predefined knowledge set  $\mathcal{K}^*$ , and 0 otherwise.

KR evaluates the model’s ability to recall and employ relevant knowledge from our collection. 
Unlike KP, which focuses on correctness, KR assesses coverage—it measures whether the model incorporates essential knowledge without missing it. This combination of complementary metrics evaluates overall reasoning quality, with (iii) \textbf{Knowledge F1} calculated as the harmonic mean of knowledge precision and recall.

\subsection{Distilling LLM Evaluator}
\label{sec:training_evaluator}
To reduce computational overhead, we train a smaller, open-weight LLM evaluator by distilling the teacher model, GPT-4o ~\citep{openai2024gpt4o}, into a more lightweight student model, Llama3-8B-Instruct ~\citep{dubey2024llama}.


The training begins by generating five diverse reasoning traces per question using a Llama3-8B-Instruct model. A teacher model then evaluates these traces against the principal knowledge set $\mathcal{K}^*$. Based on these evaluations, a student model is fine-tuned to mimic the teacher’s assessments, resulting in a cost-efficient, open-weight evaluator that preserves evaluation quality.

In our pilot experiments, the student model successfully learned to evaluate generated rationales, achieving F1 scores of 96.3 for judging whether the used knowledge is accurate and 94.7 for determining whether the principal knowledge is present in the rationale, closely matching the teacher model’s performance.\footnote{See detailed experiment in Appendix.~\ref{appendix:evaluator_train_detail}}

\section{Aligning LLM Reasoning with Knowledge-Grounded Metrics}
\label{RDPO}

\paragraph{Preliminary: Direct Preference Optimization}
The preference optimization paradigm fine-tunes language models based on human or task-specific preferences to generate more desirable outputs. A prominent method, Direct Preference Optimization (DPO)~\cite{rafailov2023direct}, trains a model $\pi_{\theta}$ to prefer response $y_w$ over $y_l$ using the objective:

$$
J(\theta) = \mathbb{E}_{(x, y_w, y_l) \sim \mathcal{D}} \log \sigma \big( r_{\theta}(x, y_w) - r_{\theta}(x, y_l) \big)
$$

with reward function
$r_{\theta}(x, q) = \beta \left[ \log \pi_{\theta}(q \mid x) - \log \pi(q \mid x) \right]$
where $\pi$ is a frozen reference model and $\beta$ a scaling factor.
The quality of the triplet dataset $\mathcal{D}$ is crucial for guiding model behavior.

Iterative Reasoning Preference Optimization (IRPO)~\cite{pang2025iterative} extends preference-based fine-tuning to reasoning tasks by iteratively  
(1) sampling multiple CoT and answer candidates per prompt, and  
(2) training the model to prefer generations with correct CoT over incorrect ones.  

\paragraph{Reasoning Preference Optimization via Knowledge-Grounded Metrics}
We further extend the utilization of our evaluation suite by integrating it into IRPO. 
As we defined, high Knowledge Precision (KP) indicates that a model correctly uses the required knowledge and minimizes hallucinations, while a high Knowledge Recall (KR) means that the model conducts an extensive exploration of the relevant knowledge. However, a correct final answer does not have to be explored through every piece of principal knowledge: sometimes more concise reasoning can suffice. Building on these insights, we incorporate KR into standard preference optimization techniques, enabling us to tune the depth of a model’s reasoning—from minimal knowledge usage to full coverage—based on specific goals such as efficiency or thoroughness.

Our approach begins by prompting LLMs to generate multiple CoT rationales for each question. 
These rationales are then categorized as preferred or dispreferred based on specific selection criteria according to the baselines discussed in the following paragraphs. 
To improve preference selection, we explore several variants by using several different strategies to build preference pairs:

\paragraph{Answer-Driven Selection} 
A simple baseline where any rationale that successfully arrives at the correct final answer is randomly designated as a preferred sample. Incorrect rationales are dispreferred, following conventional IRPO selection.

\paragraph{Assigning Knowledge Importance}
We first assign an importance weight  $w_i = |\mathcal{C}_i|$, corresponding to the cluster size of each knowledge element in the knowledge collection. These weights reflect the prevalence and significance of certain knowledge elements, allowing our evaluation to prioritize reasoning that relies on high-impact knowledge. Based on this weight, we calculate the weighted KP and KR which are used for configuring preference set.~\footnote{See detailed weighted knowledge score calculation in Appendix.~\ref{appendix:weighted_metrics}}

\paragraph{Ensuring Reasoning Factuality via weighted KP}
Among the correct rationales, we select one with the highest weighted KP to set as the preferred sample. This helps preserve factual correctness since a higher KP reflects fewer hallucinations or incorrect facts. On the other hand, we select one with the lowest weighted KP as the dispreferred sample among the incorrect rationales. 

\paragraph{Controlling Reasoning Comprehensiveness via weighted KR}
When multiple correct rationales share the same highest weighted KP, we break ties by weighted KR. Depending on our goal, we may:
\begin{itemize}
    \item \textbf{Random KR}: Select among these top-KP rationales at random, providing a baseline without KR targeting.
    \item \textbf{Maximize KR}: Choose the rationale with the \textbf{highest} weighted KR to encourage broader usage of principal knowledge.
    \item \textbf{Minimize KR}: Choose the rationale with the \textbf{lowest} weighted KR to promote more efficient reasoning while preserving correctness.
\end{itemize}

Regardless of the specific KR tie‐breaking strategy, the dispreferred samples are chosen oppositely. In cases where all rationales are correct, we continue applying the same selection logic to identify both the preferred and dispreferred samples from among the correct answers. Conversely, if all rationales are incorrect, we discard those samples entirely.

\section{Experiments}
In this section, we present two key experiments. First, we evaluate how effectively LLMs apply knowledge in their reasoning using the \textsc{PK Collection} and our proposed knowledge-grounded metrics. Then, we examine the impact of our knowledge-grounded in reasoning alignment, demonstrating how controlling KR influences the model’s reasoning process. 

\newcommand{\cmark}{\ding{51}} 
\newcommand{\xmark}{\ding{55}} 

\begin{table*}[]
\centering
\resizebox{\textwidth}{!}{%
\begin{tabular}{lcccccccccc}
\toprule
\multirow{2}{*}{\textbf{Models}} & \multirow{2}{*}{\textbf{Acc.(\%)}} & \multicolumn{3}{c}{\textbf{Knowledge Precision (\%)}} & \multicolumn{3}{c}{\textbf{Knowledge Recall (\%)}} & \multicolumn{3}{c}{\textbf{Knowledge F1 (\%)}}  \\ 
\cmidrule(lr){3-5} \cmidrule(lr){6-8} \cmidrule(lr){9-11}
  &  & \textbf{all} & \textbf{correct} & \textbf{incorrect} & \textbf{all} & \textbf{correct} & \textbf{incorrect} & \textbf{all} & \textbf{correct} & \textbf{incorrect} \\
\midrule
\textit{\textbf{Up to 10B}} &  &  &  &  &  &  &  &  &  \\ 
Llama3-8b-instruct & 64.6 & 92.4 & 98.6 & 81.1 & 83.1 & 87.0 & 76.0 & 85.7 & 91.4 & 75.5 \\
Llama3.1-8b-instruct & 70.6 & 95.0 & 99.3 & 84.7 & {\ul 89.7} & 92.5 & 82.9 & {\ul 91.3} & 95.3 & 81.6 \\
Mistral-7B-Instruct-v0.3 & 60.7 & 92.0 & 98.1 & 82.6 & 76.5 & 79.9 & 71.3 & 81.3 & 86.3 & 73.5 \\
Phi-3-mini-4k-instruct & 73.2 & 94.4 & 99.2 & 81.2 & 86.0 & 89.6 & 76.1 & 88.7 & 93.4 & 75.7 \\
Phi-3-small-8k-instruct & \textbf{79.2} & \textbf{96.2} & 99.3 & 84.3 & 87.4 & 89.9 & 77.6 & 90.5 & 93.7 & 78.2 \\
Qwen2.5-7B-Instruct & {\ul 75.1} & {\ul 95.8} & 99.3 & 85.2 & \textbf{90.2} & 92.1 & 84.5 & \textbf{92.0} & 95.1 & 82.8 \\ \hline
\textit{\textbf{Larger than 10B}} &  &  &  &  &  &  &  &  &  &  \\
Mixtral-8x7B-Instruct-v0.1 & 69.8 & 95.0 & 99.1 & 85.5 & 83.8 & 86.3 & 78.0 & 87.5 & 91.2 & 78.9 \\
Phi-3-medium-4k-instruct & {\ul 79.6} & {\ul 96.1} & 99.3 & 83.4 & {\ul 86.6} & 89.0 & 77.4 & {\ul 89.9} & 93.1 & 77.6 \\
Qwen2.5-32B-Instruct & \textbf{84.1} & \textbf{97.7} & 99.5 & 87.9 & \textbf{91.7} & 92.8 & 85.7 & \textbf{93.9} & 95.6 & 85.0 \\ \hline
\textit{\textbf{R1 distilled}} &  &  &  &  &  &  &  &  &  &  \\
DeepSeek-R1-Distill-Llama-8B & 72.9 & 92.5 & 99.0 & 75.2 & 78.6 & 82.7 & 67.5 & 83.0 & 88.6 & 67.9 \\
DeepSeek-R1-Distill-Llama-70B & \textbf{88.4} & \textbf{98.0} & 99.3 & 88.2 & \textbf{90.9} & 91.8 & 84.4 & \textbf{93.7} & 94.9 & 84.5 \\
DeepSeek-R1-Distill-Qwen-7B & 68.0 & 86.3 & 98.2 & 60.9 & 67.3 & 75.8 & 49.1 & 73.0 & 83.4 & 50.9 \\
DeepSeek-R1-Distill-Qwen-32B & {\ul 86.7} & {\ul 97.6} & 99.5 & 84.9 & {\ul 89.2} & 90.2 & 82.6 & {\ul 92.3} & 94.0 & 81.5 \\
\bottomrule
\end{tabular}%
}
\caption{Reasoning performance of various LLMs evaluated on our evaluation suite. \textbf{Bold} indicates the top score and \underline{underline} means the second best score in each group. "Correct" and "Incorrect" refer to evaluations of reasoning steps from final answers that were correct or incorrect, respectively.}
\label{tab:performance_comparison}
\end{table*}

\subsection{Experimental Setup}

\paragraph{Datasets}
We evaluate our method on the MMLU benchmark, which spans diverse subjects to assess domain-specific knowledge and reasoning.

Since MMLU does not provide a dedicated training set, we use an 8:2 train-test split per subject. For each question, we generate 32 rationales to build a candidate pool for DPO, selecting preferred samples based on the policy in Section~\ref{RDPO}. Dataset statistics are in Appendix~\ref{appendix:data_stats}.



    

\paragraph{Models}
We evaluate open-source LLMs to gauge how effectively they apply knowledge in their reasoning. We examine LLaMA-3-8B-Instruct~\cite{dubey2024llama}, Mistral-7B-Instruct-v0.3~\cite{jiang2023mistral}, Mixtral-8×7B-Instruct-v0.1~\cite{jiang2024mixtral}, as well as Qwen2.5-(7B, 32B)-Instruct~\cite{qwen2025qwen25technicalreport} and Phi‐3-(mini‐4k, small‐8k, medium‐4k)~\cite{abdin2024phi3}.
Additionally, we include distilled variants of Deepseek R1 (Llama‐8B, Llama‐70B, Qwen‐7B, Qwen‐32B)~\cite{deepseekai2025deepseekr1incentivizingreasoningcapability}.\footnote{See details for Deepseek R1 variants in appendix~\ref{appendix:deepseek}.}

\paragraph{Evaluation Metrics}
We report standard multiple-choice accuracy on the MMLU benchmark, along with our proposed metrics (Section~\ref{sec:eval_reasoning}), which assess how accurately and comprehensively each model applies the required knowledge.




\subsection{LLM Performances on Our Evaluation Suite}
Table~\ref{tab:performance_comparison} shows the overall performance of each model on our evaluation suite where all evaluations are conducted in a zero‐shot CoT setting. 
For knowledge-grounded metrics, we present the results broken down by whether the final answer is correct or incorrect.
We also observe that, for most models, knowledge-grounded metrics are lower on incorrect questions compared to correct ones. This pattern suggests that 
LLMs often produce incorrect results when they fail to recall relevant information or apply it precisely during the reasoning process.
We group the models into three categories and present the results:

\begin{table}[]

\centering
\resizebox{\columnwidth}{!}{%
\begin{tabular}{lcccc} \toprule
 & \textbf{KP} & \textbf{KR} & \textbf{F1} & \textbf{Acc.} \\ \midrule
\textit{Zero-Shot CoT} & 90.9 & 82.0 & 84.5 & 66.3  \\
\textit{DPO} &  &  &  &   \\
\quad Answer-Driven & 94.4{\footnotesize(+3.5)} & 85.9{\footnotesize(+3.9)} & 88.6{\footnotesize(+4.1)} & 70.3{\footnotesize(+4.0)}\\
\quad Random KR & 96.2{\footnotesize(+5.3)} & 87.6{\footnotesize(+5.6)} & 90.6 {\footnotesize(+6.1)} & 70.8 {\footnotesize (+4.5)}  \\
\quad Max KR & 96.5{\footnotesize(+5.6)} & 92.6{\footnotesize(+10.6)} & 93.8 {\footnotesize(+9.3)} & 70.1 {\footnotesize (+3.8)}  \\
\quad Min KR & 94.8{\footnotesize(+3.9)} & 74.4{\footnotesize(-7.6)} & 81.0 {\footnotesize (-3.5)} & 70.6{\footnotesize (+4.4)} \\ \bottomrule
\end{tabular}%
}
\caption{Performance of Llama3-8B-Instruct in a zero-shot setup and after fine-tuning with the DPO objective, both utilizing various data selection strategies on the MMLU test split. Score differences relative to \textit{Zero-Shot} are indicated in parentheses.}
\label{tab:llama_dpo}
\end{table}

\begin{table*}[t]
\centering
\scalebox{1}{
\begin{tcolorbox}[colback=gray!3, colframe=black, title=Example: Principal Knowledge Collection in College Mathematics]

\begingroup
\setlength{\baselineskip}{0.9\baselineskip}

\textbf{Question.} Water drips out of a hole at the vertex of an upside-down cone at a rate of 3 cm$^3$ per minute. The cone’s height and radius are 2 cm and 1 cm, respectively. At what rate does the height of the water change when the water level is half a centimeter below the top of the cone? The volume of a cone is $V = (\pi/3)r^2h$. \\
\vspace{0.5em}
\begin{tabular}{l}
(A) $-48/\pi$ cm/min \\
(B) $-4/\pi$ cm/min \\
(C) $-8/3\pi$ cm/min \\
(D) $-16/3\pi$ cm/min \\
\end{tabular}

\vspace{0.8em}
\textbf{Principal Knowledge}
\begin{itemize}[itemsep=-2pt, topsep=0pt]
    \item The chain rule is used to relate the rates of change of volume, radius, and height.
    \item The relationship between the radius and height of the water in the cone is proportional, given by similar triangles.
    \item The volume of a cone is given by the formula $V = (\pi/3)r^2h$.
    \item The negative sign in the rate of change indicates that the height of the water is decreasing over time.
    \item Implicit differentiation is used to find the rate of change of the height of the water with respect to time.
    \item The concept of related rates involves using derivatives to find the rate of change of one quantity with respect to another.
    \item The rate of change of volume with respect to time is $3\,\text{cm}^3/\text{min}$.
    \item The derivative of the volume of the cone with respect to height ($dV/dh$) is $(\pi/3)r^2$, where $r$ is the radius of the water at the given height.
    \item The rate of change of the height of the water ($dh/dt$) can be found using the chain rule: $dV/dt = (dV/dh)(dh/dt)$.
\end{itemize}
\endgroup
\end{tcolorbox}
}
\captionsetup{skip=8pt, labelfont=bf}
\caption{An example of \textsc{PK Collection} for a MMLU College Mathematics problem. This structured list highlights all required mathematical concepts, such as the chain rule, implicit differentiation, and geometric relationships, which are essential to evaluating model understanding in symbolic reasoning tasks.}
\label{table:pk_collection}
\end{table*}

\textbf{Up to 10B} Among smaller models, Phi‐3‐small‐8k‐instruct stands out with the highest accuracy (79.2\%) and consistently high KP (96.2\%) and KR (87.4\%), indicating both correct and comprehensive knowledge usage.

\textbf{Larger than 10B} In this group, Qwen2.5‐32B‐Instruct achieves the best KP (97.7\%) and KR (91.7\%) and excels in accuracy (84.1\%), reflecting strong reasoning in terms of factuality and depth.

\textbf{R1 Distilled} DeepSeek R1 Distilled models generally improve over their non-distilled counterparts. DeepSeek‐R1‐Distill‐Llama‐70B, for instance, attains the highest overall accuracy (88.4\%) and maintains robust KP and KR, suggesting that the specialized “deep thinking” strategy meaningfully enhances both correctness and effective knowledge application. 

Appendix~\ref{appendix:case_study_distill} compares R1 distilled and non-distilled models in detail.

Another insight is that differences in knowledge recall across models are notably greater than differences in precision. Lower recall indicates that models often fail by not retrieving relevant knowledge, either omitting essential units or introducing irrelevant knowledge, leading to incorrect answers. Thus, knowledge recall effectively captures how broadly and appropriately a model applies knowledge during reasoning.


\subsection{Further Enhancement from Reasoning Preference Optimization}
Table~\ref{tab:llama_dpo} presents the results of the enhancement of Reasoning Preference Optimization incorporating our knowledge-grounded metric (\S \ref{RDPO}), demonstrating how different selection strategies influence model performance. 
Answer-driven selection, which follows the conventional IRPO approach, improves KP, KR, and accuracy. While effective, it does not explicitly control for knowledge scores. In contrast, our selection strategies prioritize rationales with maximum KP, ensuring less hallucinated reasoning and consistently leading to higher KP across all models.

Random KR improves KP, KR, and accuracy over the zero-shot CoT and answer-driven selection method, validating reasoning preference optimization incorporating KP. Max KR selection leads to +10.0\%p KR and +3.8\%p final-answer accuracy, showing that broader knowledge application enhances problem-solving. Conversely, Min KR maintains a high KP while reducing KR by -8.2\%p, yet still improves accuracy (+4.4\%p), suggesting that concise, essential reasoning can be as effective as extensive usage of knowledge.

As shown in Table~\ref{table:minmax}, by picking responses with either the lowest or highest KR responses, we can guide the model toward more concise reasoning or more comprehensive reasoning, respectively. Appendix~\ref{appendix:case_study_minmax} explains a detailed case study illustrating how this selection strategy affects reasoning quality and output characteristics.

\section{Example and Case Study}
This section presents illustrative example and case study to demonstrate the application of our evaluation framework. 
\subsection{Example of \textsc{PK Collection}}
\label{appendix:pk_collection_example}

We provide an illustrative example of how we extract PK Units from an MMLU college mathematics problem involving related rates. As shown in Table~\ref{table:pk_collection}, the problem requires an understanding of the geometric relationship between radius and height (via similar triangles), the application of the chain rule, and the technique of implicit differentiation.
Each atomic knowledge encapsulates a specific concept or operation required for solving the problem. By decomposing the solution path into these discrete components, we can systematically evaluate whether a model’s reasoning includes, omits, or misapplies any part of the necessary conceptual foundation.
This example demonstrates how \textsc{PK Collection} enables fine-grained, interpretable analysis of model reasoning in symbolic and quantitative domains.

\subsection{Comparision of R1-Distilled and Non-Distilled models}
\label{appendix:case_study_distill}

We present a case study comparing two models with the same parameter size: DeepSeek-R1-Distilled-Llama-8B (distilled) and Llama3.1-8B-Instruct (non-distilled), evaluated on a factual multiple-choice science question involving nuclear fission.
Table~\ref{table:distill_comparision} shows that the distilled model provided a concise but accurate rationale, applying only 3 out of 7 PK units (KR = 42.9\%) with 100\% precision (KP = 100\%), leading to the correct final answer. In contrast, the non-distilled model applied all 7 PK units (KR = 100\%, KP = 100\%) but still failed to select the correct answer due to misjudgment at the conclusion stage. This highlights how concise reasoning with correct knowledge can outperform broader yet flawed reasoning.

\section{Analysis}

\paragraph{Are the Principal Knowledge in the \textsc{PK Collection} Factual and Relevant?}
To validate the factuality of the \textsc{PK Collection}, we employed GPT-4o with OpenAI’s WebSearch tool API to conduct web-based fact-checking. We performed binary classification (True/False) for each of 4k PK units across 512 MMLU questions. This process yielded an overall accuracy of \textbf{94.1\%}. Among the 232 PK units marked as False, we manually reviewed a random sample of 100. As shown in Table~\ref{table:human_eval}, only 33\% were genuinely factually incorrect, while the remaining cases were due to various factors such as ambiguity, model deficiency, or overly strict judgments. This suggests that the actual number of incorrect PK units is much smaller, and that the \textsc{PK Collection} constitutes a highly accurate set.

For relevance, we adopted a scoring rubric inspired by the Prometheus approach~\citep{kim2024prometheus}, rating the relevance of each PK unit to its corresponding question on a 1–5 scale. Using GPT-4o, we assessed the same 4k PK units, yielding an average relevance score of \textbf{4.0}.

Together, these validations confirm that the \textsc{PK Collection} maintains high quality in terms of both factual accuracy and relevance. \footnote{See the detailed validation procedure in Appendix~\ref{appendix:validation_details}}


\paragraph{Does Providing Misused or Unused Knowledge Improve Accuracy?}
We investigate the quality of principal knowledge in our collection by incorporating it in the prompt, adopting Llama3‐8B‐Instruct.
We pick error cases from the benchmark and prompt LLM to answer the questions based on 
one of four types of knowledge elements: (1) random--a randomly chosen knowledge element, (2) correct-- knowledge that was correctly used in the original reasoning, (3) incorrect--knowledge that was previously misapplied, or (4) all-- 
knowledge that was either misapplied or not utilized (unapplied) in the model’s original reasoning.

\begin{figure}[t!]
\centering
\includegraphics[width=\columnwidth]{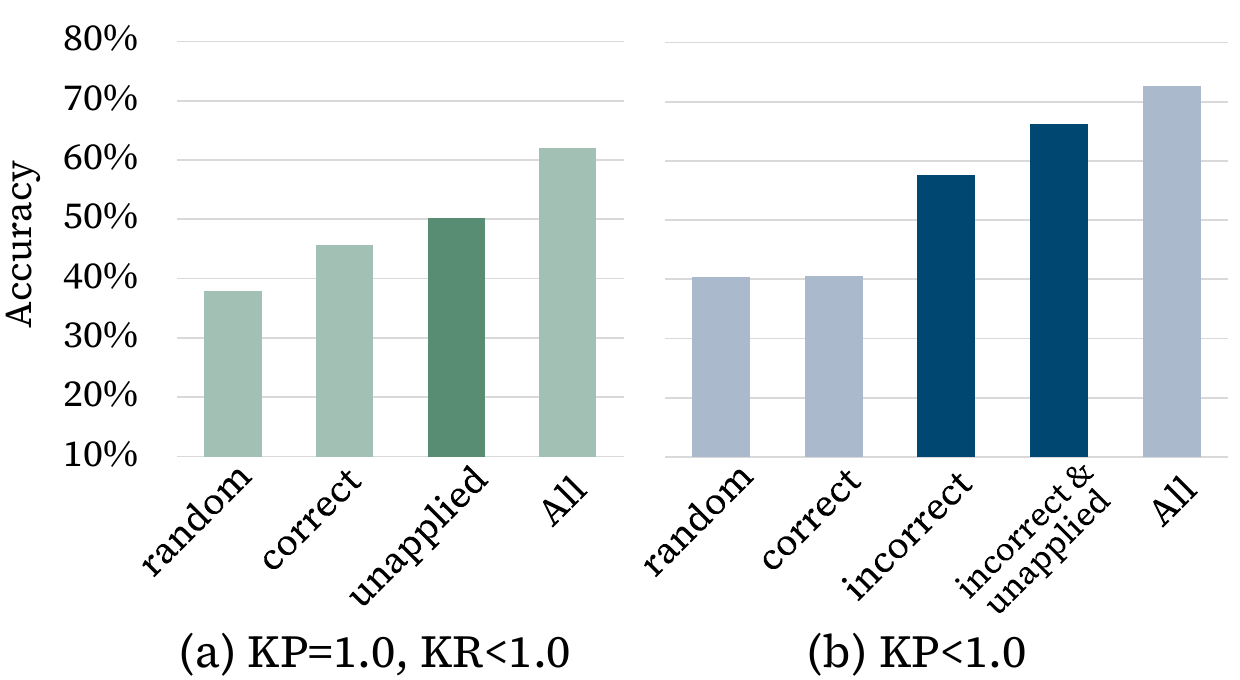}
\caption{
Accuracy improvement when incorrect questions are retried with additional knowledge. (a) Cases where all applied knowledge was correct but some necessary knowledge was missing (KP=1.0, KR < 1). (b) Cases where some knowledge was misapplied (KP < 1).}
\label{fig:knowledge acccuracy bar plot}
\vspace{-0.3cm}
\end{figure}

Figure~\ref{fig:knowledge acccuracy bar plot} (a) shows cases where all applied knowledge was correct but some required knowledge was missing (KP=1, KR < 1). As the chart shows, providing these unapplied knowledge elements significantly boosts accuracy.
This suggests that while the model's reasoning was partially correct, it did not fully leverage sufficient relevant knowledge; once the missing knowledge is provided, the model can arrive at the correct answer.


Figure~\ref{fig:knowledge acccuracy bar plot} (b) illustrates the cases where some knowledge in its original reasoning was incorrectly used (KP < 1). Under these conditions, giving the misapplied knowledge can rectify the model’s misconception, substantially raising the accuracy. When both unapplied and misapplied knowledge are provided, we observe accuracy surging to 66.2\%, indicating that addressing sufficient and correct knowledge usage can transform previously erroneous solutions into correct ones.

These results highlight how our evaluation suite effectively diagnoses model weaknesses, revealing whether failures arise from missing or misapplied knowledge. Furthermore, it provides insights into targeted interventions, guiding strategies to improve the model’s reasoning by supplementing critical knowledge gaps.


\paragraph{Can Alignment with KR Reduce Token Consumption?}

\begin{figure}[t!]
\centering
\includegraphics[width=\columnwidth]{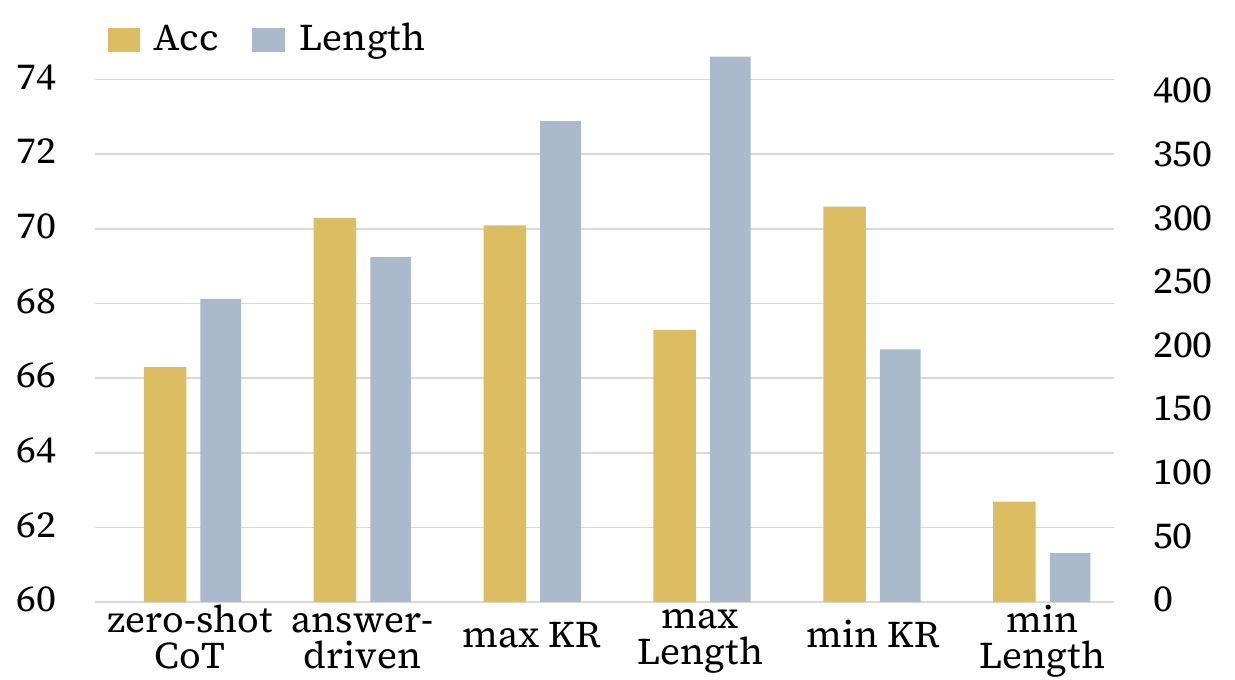}
\caption{
Comparison of accuracy and token length in various data selection settings. Details about length based selection method are described in Appendix~\ref{appendix:length_base}}
\label{fig:token length comparison plot}
\vspace{-0.3cm}
\end{figure}

We investigate whether our KR-based approach influences the overall length of the model’s reasoning. 
As shown in Figure~\ref{fig:token length comparison plot}, answer-driven setups improve accuracy compared to the zero-shot CoT but tend to produce longer rationales—reflecting more extensive knowledge exploration encouraged by reasoning preference optimization. Similarly, explicitly maximizing KR further raises accuracy while significantly increasing token usage.

A natural question is whether simply choosing the “longest” correct rationales yields the same benefit as max KR. While the max length approach indeed generates more tokens, it does not achieve the same accuracy gains, indicating that sheer verbosity does not necessarily equate to broader knowledge usage. Conversely, minimizing KR effectively reduces token length while still improving accuracy relative to zero-shot CoT. By contrast, the min length approach—selecting the shortest correct rationales—drastically reduces token usage but also causes a sharp drop in final accuracy. These comparisons underscore that rationale length alone is not a reliable proxy for knowledge coverage or correctness; KR-based selection strikes a more meaningful balance between thorough exploration and concise, accurate reasoning.~\footnote{See the performance of length-based approach in Table~\ref{length_based_performance}}


\section{Conclusion} 
We introduce a knowledge-grounded evaluation suite that assesses LLMs' ability to accurately recall and apply essential knowledge in reasoning. By constructing the Principal Knowledge Collection, we establish a large-scale resource of atomic knowledge, which serves as the foundation for our knowledge-grounded metrics—precision, recall, and F1—measuring both correctness and completeness in reasoning. Our lightweight evaluator LLM replicates a proprietary teacher model’s assessment while reducing computational costs. Experimental results show that integrating these metrics into preference optimization techniques enhances model performance, enabling controlled reasoning efficiency. These findings highlight the effectiveness of our evaluation suite in improving the transparency and reliability of LLM reasoning.

\section*{Limitations}

Even top-performing proprietary models like GPT-4o and leading open-source LLMs can occasionally make errors when extracting atomic knowledge necessary for solving a given question. Since our evaluation process inherently relies on LLMs, errors can arise, and the computational cost of large-scale evaluation is significant. To mitigate these issues, we leverage four high-performing LLMs and apply clustering techniques to refine the knowledge set, reducing inconsistencies in data construction. Additionally, we generate evaluation data using GPT-4o, a strong general-purpose model, and train a lightweight evaluator model to replicate its assessment capabilities. This approach significantly lowers computational costs while maintaining comparable evaluation performance. However, despite these efforts, some inaccuracies may still persist in our data construction process.

\section*{Acknowledgments}
We thank Taewhoo Lee and Minbyeol Jung for the valuable feedback on our work.
This work was supported in part by the National Research Foundation of Korea [NRF-2023R1A2C3004176], the Ministry of Health \& Welfare, Republic of Korea [HR20C002103], the Ministry of Science and ICT (MSIT) [RS-2023-00262002], and the ICT Creative Consilience program through the Institute of Information \& Communications Technology Planning \& Evaluation (IITP) grant funded by the MSIT [IITP-2025-RS-2020-II201819].

\section*{Ethics Statement}
Our evaluation process can incur significant environmental costs due to its computationally intensive nature. 
To reduce this impact, we fine-tune a lightweight evaluator model to lower computational expenses.
Additionally, a potential concern is that the generated dataset may include biases, such as stereotypes related to race or gender. While no major issues have been reported in the creation of question-answering datasets to our knowledge, incorporating robust training or validation methods to mitigate such biases would be beneficial.

\bibliography{custom}

\appendix
\clearpage
\renewcommand{\thetable}{\Alph{table}}
\setcounter{table}{0}
\clearpage
\section*{Appendix}
\
\appendix
\label{sec:appendix}
\section{Data Statistics}
\label{appendix:data_stats}

\begin{figure}[H]
\centering

\includegraphics[width=\columnwidth]{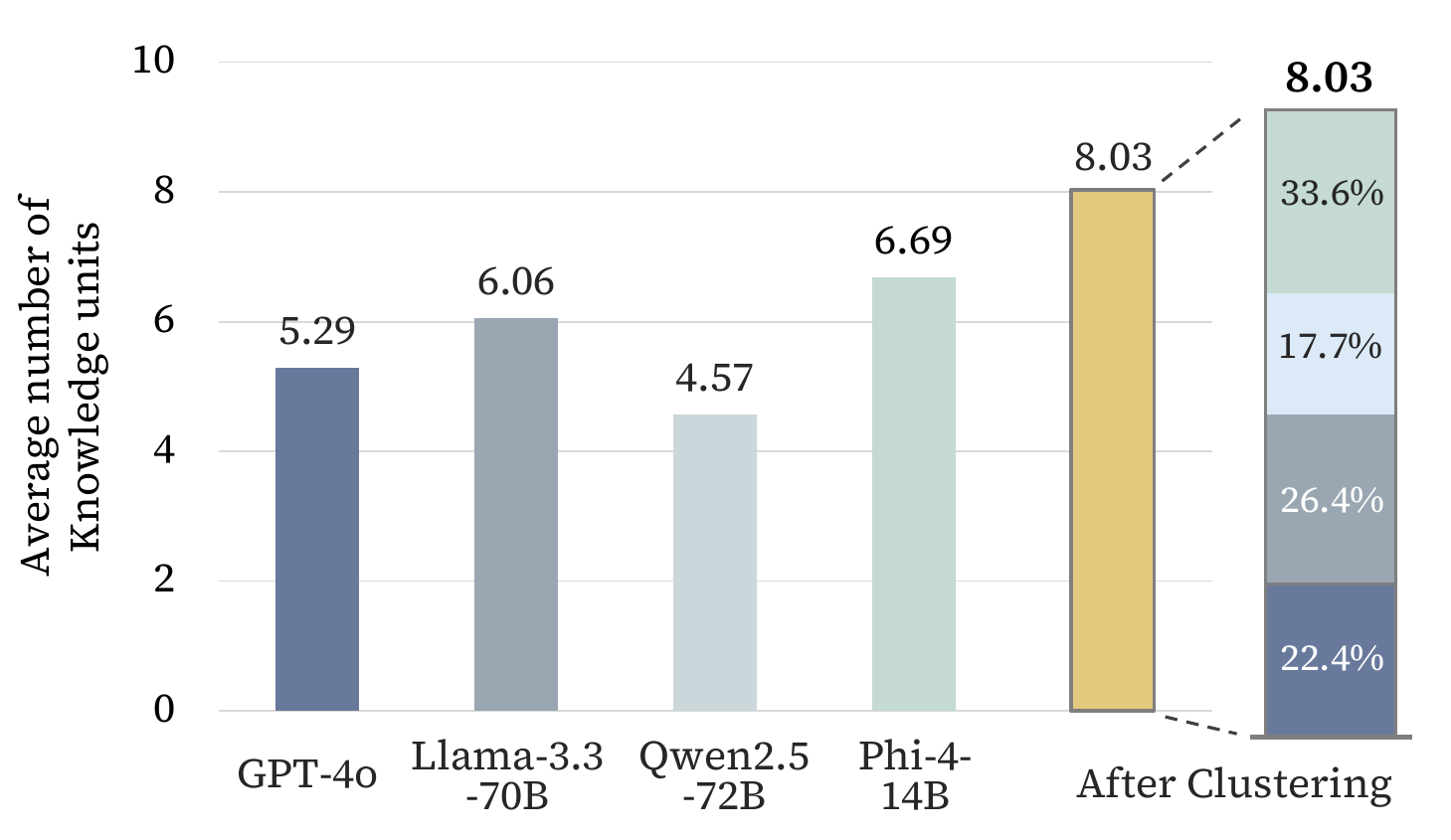}

\vspace{-0.2cm}
\caption{
The average number of atomic knowledge generated by each LLM before and after clustering. We define principal knowledge as the closest knowledge to the centroid of each cluster and show the proportions of this knowledge coming from each model.
}
\label{fig:knowledge_avg}
\vspace{-0.3cm}
\end{figure}

\begin{figure}[H]
\centering

\includegraphics[width=\columnwidth]{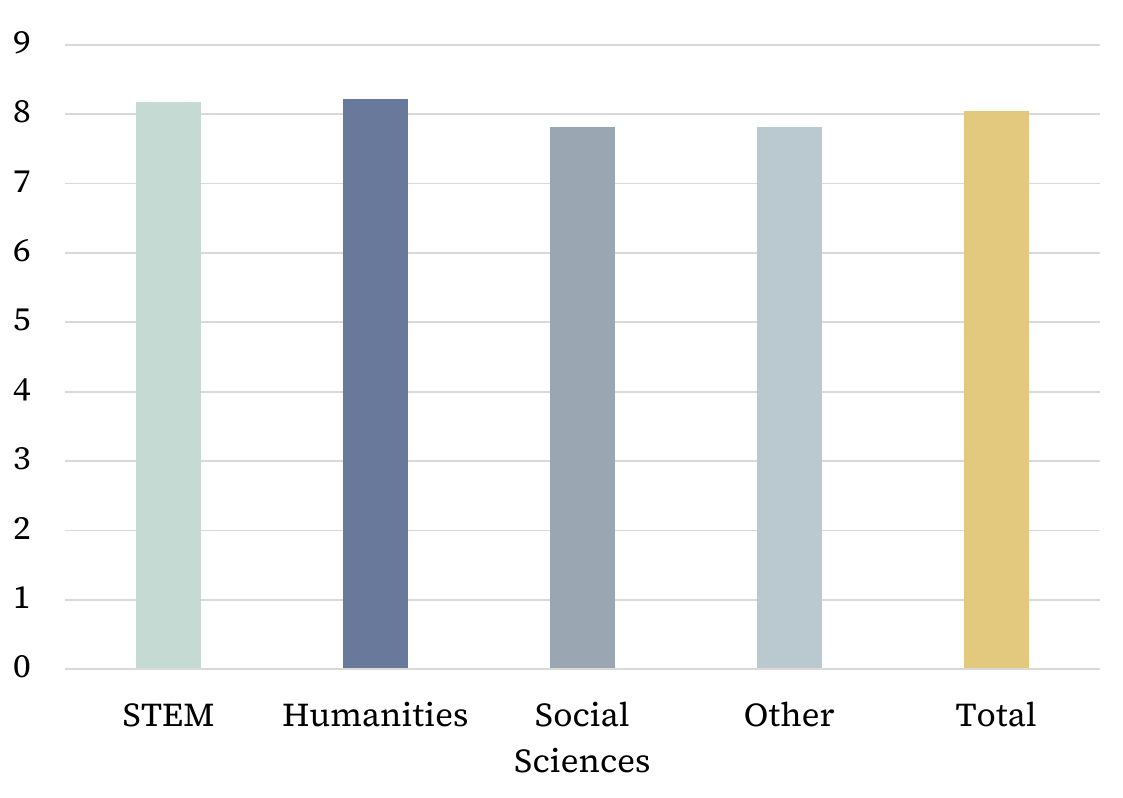}

\vspace{-0.5cm}
\caption{
The average number of Principal Knowledge Collection (category-wise).
}

\label{fig:knowledge_avg_categrywise}
\vspace{-0.3cm}
\end{figure}

Statistics of \textsc{PK Collection} and MMLU is shown in figure~\ref{fig:knowledge_avg}, figure~\ref{fig:knowledge_avg_categrywise}  and Table~\ref{tab:mmlu_statistics}.


Table~\ref{tab:subjects} also provides the category-wise classification of MMLU subjects used in our experiments.

\begin{table}[H]
\resizebox{\columnwidth}{!}{%
\begin{tabular}{lccc} \toprule
 & \textbf{\# of questions} & \textbf{\# of knowledge} & \textbf{Avg. \# of knowledge} \\ \midrule
Total & 14043 & 112780 & 8.03 \\ \midrule
Train & 11216 & 90027 & 8.03 \\
Test & 2827 & 22753 & 8.05 \\ \bottomrule
\end{tabular}%
}
\caption{Statistics of MMLU Question and KP Collection}
\label{tab:mmlu_statistics}
\end{table}

\section{Training Details of Evaluator}
\label{appendix:evaluator_train_detail}

We train our evaluator using four Nvidia H100 GPUs, each with 80GB of memory. Our implementation is based on PyTorch~\citep{paszke2019pytorchimperativestylehighperformance} and HuggingFace~\citep{wolf2020huggingfacestransformersstateoftheartnatural}. The training follows a supervised fine-tuning approach as outlined in the published alignment handbook~\citep{Tunstall_The_Alignment_Handbook}. We optimize the model using the Adam optimizer~\citep{kingma2017adammethodstochasticoptimization} with a learning rate of 2e-6, a batch size of 64, and a warm-up ratio of 0.1 over 5 epochs. For inference, we employ a greedy decoding strategy with a temperature of 0 and top\_p set to 1.0 across all experiments.\begin{enumerate}
    \item \textbf{Data Collection via GPT-4o Annotations.} We begin with an 80\% subset of MMLU questions (denoted \(\mathcal{Q}_{\text{train}}\)). For each question \(q\), we sample five reasonings using a Llama3-8B-Instruct model, resulting in five distinct reasonings \(r_j\). We then provide each \(\langle q,\;r_j\rangle\) pair, along with its associated problem-solving knowledge set \(\mathcal{K}^*(q)\), to GPT-4o for annotation. Concretely, GPT-4o labels for each knowledge element \(\hat{k}_i\in\mathcal{K}^*(q)\) whether it is ``\textit{unapplied},'' ``\textit{incorrect},'' or ``\textit{correct}'' in \(r_j\). These annotations serve as our training data.
    
    \item \textbf{Training the Evaluator.} Next, we fine-tune a Llama3-8B-Instruct model to mimic these GPT-4o labels. The training objective is to predict, for each \(\hat{k}_i\), whether it is used correctly, used incorrectly, or omitted in the chain-of-thought \(r_j\). Formally, we train an evaluator \(E\) such that:
    \[
    E\bigl(r_j,\;\hat{k}_i\bigr)\;\approx\; \text{Label}_{\text{GPT-4o}}(r_j,\;\hat{k}_i).
    \]

    \item \textbf{Evaluator Testing.}
    After training, we collect newly sampled CoT reasonings for \emph{both} the training (\(80\%\)) and testing (\(20\%\)) MMLU questions. In this stage, each question has a fresh reasoning \(r\), and our evaluator \(E\) assigns labels to each \(\hat{k}_i \in \mathcal{K}^*(q)\). We then compare these labels with GPT-4o’s annotations.
    
    \item \textbf{Evaluator Performance.} We measure two F1 scores:

 \(\textbf{Application F1}\): The agreement in whether each knowledge element \(\hat{k}_i\) was identified as \emph{applied} (correct or incorrect) vs.~\emph{unapplied}.
 
\(\textbf{Correctness F1}\): The agreement in distinguishing \textit{correct} usage from \textit{incorrect} usage, conditioned on those elements identified as applied.

\end{enumerate}

Empirically, we find that our evaluator achieves an \(\textbf{Application F1}\) of \(\mathbf{96.26}\) and a \(\textbf{Correctness F1}\) of \(\mathbf{94.69}\) on \(\mathcal{Q}_{\text{test}}\). These results indicate that a carefully fine-tuned model of moderate size can robustly replicate GPT-4o’s knowledge application judgments at significantly lower computational cost. 

\section{Details of Experiment}
For sampling, we use a temperature of 1.0. In other cases, we employ greedy decoding. The maximum token limit is set to 8192.

\subsection{DPO Training Details}
\label{appendix:DPO_training}
We use four Nvidia H100 GPUs (80GB memory each) for DPO training. Our code is implemented in PyTorch~\citep{paszke2019pytorchimperativestylehighperformance} and HuggingFace~\citep{wolf2020huggingfacestransformersstateoftheartnatural}, with DeepSpeed Stage 3~\citep{rajbhandari2020zero} for multi-GPU training and FlashAttention~\citep{dao2022flashattention} for efficiency. The learning rate is set to 5e-7, and we use a $\beta$ value of 0.01 for the DPO objective.

\subsection{Knowledge Importance Weighted Precision \& Recall}
\label{appendix:weighted_metrics}

\[
\mathrm{WKP}(r) = \frac{1}{|\hat{\mathcal{K}}|} 
\sum_{i=1}^{|\hat{\mathcal{K}}|} w_i \, \mathbb{I}(\text{Eval}(\hat{k}_i, \mathcal{K}^*))
\]
\[
\text{where} \quad w_i = |\mathcal{C}_i|.
\]
$$
\mathrm{WKR}(r) = \frac{1}{|\mathcal{K}^*|} 
\sum_{i=1}^{|\hat{\mathcal{K}}|} w_i \, \mathbb{I}(\text{Eval}(\hat{k}_i, \mathcal{K}^*))
$$
$$
\text{where} \quad w_i = |\mathcal{C}_i|.
$$
\subsection{Tie-breaking rules}
\label{appendix:tie-breaking}
In case of a tie, we apply tie-breaking rules and denote the results as follows:
\begin{itemize}[itemsep=0pt, topsep=3pt]
    \item Random: Select a random candidate.
    \item KR-Based: Choose the candidate with the highest (KR-max) or lowest KR (KR-min).
    \item Length-Based: Choose the longest (length-max) or shortest response (length-min).
\end{itemize}
Among those that have the correct output, we label this chosen rationale as \textit{Preferred}. Conversely, from the set of incorrect rationales, we pick the one with the lowest Knowledge Precision using the opposite tie‐breaking rule and label it as \textit{Dispreferred}. 

\subsection{Deepseek R1}
\label{appendix:deepseek}
Deepseek-R1 models~\cite{deepseekai2025deepseekr1incentivizingreasoningcapability} incorporate a “thinking process” with <think> tags to encourage reflective reasoning. However, as this process can yield excessively lengthy reasoning with a lot of redundancies, we do not evaluate the content within the <think> tags, focusing on the final rationale that follows. The number and proportion of omitted samples due to length constraints are presented in Table ~\ref{deepseek_statistics}

\subsection{Length Based Data Selection}
\label{appendix:length_base}
For constructing a dataset for length-based preference optimization, we apply the following data selection strategies:  
\begin{itemize}[itemsep=0pt, topsep=3pt]
    \item Min Length: Among correct rationales, we select the one with the shortest token length as the preferred sample. To avoid extreme outliers, we set a minimum threshold of 100 tokens, ensuring that rationales shorter than this are not chosen. Conversely, for dispreferred samples, we select the longest incorrect rationale. 
    \item Max Length: This follows the opposite strategy, selecting the longest correct rationale as the preferred sample, while the shortest incorrect rationale is chosen as dispreferred.
\end{itemize}

\section{Comparison of Responses with Minimal vs. Maximal KR}
\label{appendix:case_study_minmax}
We present a case study comparing two model responses for the same question, selected using our sampling strategy: one with the lowest Knowledge Recall (KR-Min) and another with the highest (KR-Max). Table~\ref{table:minmax} presents the rationale generated for a question about John Maynard Keynes.
In the KR-Min example, the model utilized only 3 out of 8 PK units (KR = 37.5\%) but applied them with perfect precision (KP = 100\%) and successfully reached the correct final answer. The reasoning was concise, omitting broader context but leveraging only essential facts.
In contrast, the KR-Max response included 7 out of 8 PK units (KR = 75\%) and likewise achieved KP = 100\%. The model demonstrated comprehensive coverage of relevant knowledge, offering a rich and detailed rationale. \\
Both responses were correct, but differed in length and breadth of invoked knowledge.
This case illustrates how selecting responses by KR enables flexible trade-offs between conciseness and comprehensiveness. Our DPO sampling framework can control the reasoning style by modulating KR, while always maintaining high factual precision through KP.

\section{Validation Details}
\label{appendix:validation_details}
To ensure the quality of our principal knowledge (PK) units, we conducted a two-part validation focusing on \textit{factuality} and \textit{relevance}. Specifically, we validated each of the 4,000 PK units extracted from 512 MMLU multiple-choice questions using both automated and manual approaches. The following subsections describe the evaluation setup for each criterion in detail.

\subsection{Factuality Validation}
To validate the factual accuracy of each PK unit, we utilized GPT-4o with a web-search plugin API. The model was prompted to assess factual correctness based solely on verifiable information, without considering relevance or usefulness. The exact prompting format used is shown in Table~\ref{table:prompt_factuality}.
Each PK unit was evaluated individually. If a statement was judged to be factually incorrect (i.e., marked \texttt{False}), we conducted a secondary manual inspection to assess whether the GPT's judgment aligned with human reasoning. For this, we used the following error taxonomy:
\begin{itemize}[itemsep=0pt, topsep=3pt]
    \item \textbf{True Negative (TN)}: The PK unit was objectively and verifiably false.
    \item \textbf{Omission Bias}: Some factual elements were missing, but the omission did not directly render the statement false. GPT likely judged it incorrect due to missing context.
    \item \textbf{Ambiguous Fact}: The statement was interpretable as either true or false depending on the wording or perspective.
    \item \textbf{Overly Strict Evaluation}: Technically imprecise but generally acceptable or true under typical usage; GPT judged it false by overly strict standards.
    \item \textbf{Model Knowledge Deficiency}: The statement was factually correct, but GPT lacked domain knowledge to identify it as such.
\end{itemize}
The breakdown of false-labeled PK units by these categories is shown in Table~\ref{table:human_eval}. Notably, among the ~6\% of PK units initially labeled as false, only one-third (approximately 2\% of all PKs) were truly factually incorrect upon human re-assessment. 

\subsection{Relevance Validation}
To validate how helpful each PK unit is for answering the corresponding question, we employed GPT-4o as an evaluator using a structured scoring rubric inspired by the Prometheus framework. The prompt used for this evaluation is shown in Table~\ref{table:prompt_relevance}.
Each PK unit was assigned a relevance score ranging from 1 to 5 based on whether it represented essential, helpful, or peripheral information relative to the target question and answer. 
Across a sample of 4,000 PK units, the average relevance score was 4.0, indicating that the majority of extracted knowledge was not only factually grounded but also highly aligned with the core reasoning required to solve the task.
\newpage
\begin{table*}[]
\small
\centering
\begin{tabular}{lccccc}
\toprule
\textbf{Models}                                                                                                           & \textbf{acc} & \textbf{\begin{tabular}[c]{@{}c@{}}\# of tokens\\ (correct)\end{tabular}} & \textbf{\begin{tabular}[c]{@{}c@{}}\# of tokens\\ (incorrect)\end{tabular}} & \textbf{KP} & \textbf{KR} \\ \midrule
Qwen2.5-32B-instruct                                                                                                      & 84.1         & 285.2                                                                     & 331.2                                                                       & 97.7        & 91.7        \\
\begin{tabular}[c]{@{}l@{}}DeepSeek-R1-Distill-Qwen-32B\\ (Rationale after \textless{}think\textgreater tag)\end{tabular} & 86.7         & 181.9                                                                     & 236.9                                                                       & 97.6        & 89.2        \\ \bottomrule
\end{tabular}
\captionof{table}{Comparison between Qwen2.5-32B-instruct and DeepSeek-R1-Distill-Qwen-32B}
\end{table*} 
\begin{table*}[t]
\centering
\scalebox{1}{
\begin{tcolorbox}[colback=gray!3, colframe=black, title=Case Study: Comparing R1-Distilled vs. Non-Distilled Models]
Below is an example where we compare responses from \textbf{DeepSeek-R1-Distilled-Llama-8B} and \textbf{Llama3.1-8B-Instruct} on the same question.\\
\begingroup
\setlength{\baselineskip}{0.9\baselineskip}
\vspace{0.2em}
\textbf{Question.} 
A common nuclear fission reactor \\
\begin{tabular}{ll}
\quad (A) heats water & \hspace{7em} (B) generates electricity directly \\
\quad (C) gets energy from nothing &  \hspace{7em} (D) is a major polluter of the atmosphere
\end{tabular}
\vspace{0.2em}

\textbf{Principal Knowledge}
\begin{itemize}[itemsep=-5pt, topsep=0pt]
    \item The steam produced in a nuclear reactor is used to drive turbines that generate electricity.
    \item Nuclear fission reactors do not generate electricity directly; they require an intermediate step of heating water to produce steam.
    \item The heat produced in a nuclear reactor is used to convert water into steam.
    \item Nuclear fission reactors are considered low polluters of the atmosphere compared to fossil fuel-based power plants, as they do not emit greenhouse gases during operation.
    \item The principle of conservation of energy states that energy cannot be created or destroyed, only converted from one form to another.
    \item Nuclear fission reactors produce heat through the fission of atomic nuclei.
    \item Nuclear fission reactors do not create energy from nothing; they convert nuclear energy into thermal energy.
\end{itemize}
\vspace{0.2em}
\textbf{DeepSeek-R1-Distilled-Llama-8B Response (70 tokens):} \\
\hspace*{0.2em} A common nuclear fission reactor heats water as part of its operation, serving as a coolant and facilitating the production of steam to generate electricity through turbines. The reactor itself does not directly generate electricity but plays a crucial role in the process by producing heat. \\
\textbf{Answer:} A \\
\textbf{Correctly Applied Knowledge:} (1), (2), (3) \\
\textbf{KP:} 100\% \hspace{5em} \textbf{KR:} 42.9\% (3 out of 7) \\
\textbf{Final Answer: Correct} 

\vspace{0.5em} 

\textbf{Llama3.1-8B-Instruct Response (524 tokens, omitted for clarity):} \\
\hspace*{0.2em} To answer this question, let's break it down step by step:
1. Understanding Nuclear Fission: Nuclear fission is a process where an atomic nucleus splits into two or more smaller nuclei, releasing a significant amount of energy in the process. This process is the basis for nuclear power generation.\\
\centerline{\textit{(omission)}} 
5. Direct Energy Generation: The question mentions "generates electricity directly." While the fission process itself does not directly generate electricity, the heat from fission is used to generate steam, which then drives turbines to produce electricity. So, while the process is indirect, it's more accurate to say that the heat from fission is used to generate electricity.
Thus, the best answer is B (generates electricity directly) because it's the closest to the truth.\\
Considering these points:\\
- A. heats water: This is true, as the heat from fission is used to produce steam, which then drives turbines.\\
- B. generates electricity directly: This is somewhat misleading, as the process is indirect, but it's the closest to the truth among the options.\\
- C. gets energy from nothing: This is incorrect, as the energy comes from the fission process, which involves the splitting of atomic nuclei.\\
- D. is a major polluter of the atmosphere: This is incorrect, as the primary pollutants from nuclear reactors are radioactive materials, not greenhouse gases.\\
\textbf{Answer:} B \\
\textbf{Correctly Applied Knowledge:} (1), (2), (3), (4), (5), (6), (7) \\
\textbf{KP:} 100\% \hspace{5em} \textbf{KR:} 100\% \\
\textbf{Final Answer: Incorrect}
\endgroup
\end{tcolorbox}
}
\captionsetup{skip=8pt, labelfont=bf}
\caption{Case study comparing distilled(DeepSeek-R1-Distilled-Llama-8B) and non-distilled(Llama3.1-8B-Instruct) models under the same parameter setting.}
\label{table:distill_comparision}
\end{table*}

\begin{table*}[t]
\centering
\scalebox{1}{
\begin{tcolorbox}[colback=gray!3, colframe=black, title=Case Study: Comparing KR Min vs. KR Max]
Below is an example where we compare two responses (minimal vs. maximal KR) regarding a question about Keynes.
\begingroup
\setlength{\baselineskip}{0.9\baselineskip}
\vspace{0.4em}

\textbf{Question.} What best describes Keynes? \\
\hspace*{1em} (A) advanced Classical economic theory by making several refinements\\
\hspace*{1em} (B) showed how Say's Law operated in capitalist economies\\
\hspace*{1em} (C) was a great American economist\\
\hspace*{1em} (D) explained the cause of and cure for the Great Depression

\vspace{0.5em}
\textbf{Principal Knowledge}
\begin{itemize}[itemsep=-6pt, topsep=0pt]
    \item Keynes proposed government intervention through fiscal policy as a means to manage economic cycles and stimulate demand.
    \item John Maynard Keynes was a British economist known for his revolutionary ideas in macroeconomics.
    \item Keynes is known for his analysis of the causes and potential solutions for economic depressions, particularly the Great Depression.
    \item Keynesian economics emerged as a response to the Great Depression, challenging classical economic theories.
    \item Keynes argued that insufficient aggregate demand was a primary cause of the Great Depression and proposed government spending as a solution to stimulate the economy.
    \item The Great Depression was a severe worldwide economic downturn that began in 1929 and lasted until the late 1930s.
    \item Keynes's work, particularly \textit{The General Theory of Employment, Interest, and Money}, provided a framework for understanding economic recessions and depressions.
    \item Say's Law posits that supply creates its own demand, a concept Keynes critiqued in his work.
\end{itemize}

\vspace{0.5em}
\textbf{KR-Min Chosen (139 tokens):} \\
\hspace*{0.2em} Let's break it down step by step! 
Keynes was a British economist, not American, so option C is incorrect. Say's Law is a Classical economic concept that states that "supply creates its own demand", so option B is incorrect. Keynes actually disagreed with this idea. Classical economic theory implies a stable and self-correcting economy, which is opposite to Keynes' views, so option A is incorrect. Therefore, the correct answer must be option D: "explained the cause of and cure for the Great Depression".

\textbf{Answer:} D \\
\textbf{Correctly Applied Knowledge:} (2), (4), (8) \\
\textbf{KP:} 100\% \hspace{5em} \textbf{KR:} 37.5\% (3 out of 8) \\
\textbf{Final Answer: Correct}

\vspace{0.5em}
\textbf{KR-Max Chosen (281 tokens):} \\
\hspace*{0.2em} Let's break down the question step by step.
The question is about Keynes, so we need to think about what Keynes is known for. Keynes was a British economist, not American, so option C is incorrect. Say's Law is a concept related to classical economics, and Keynes actually disagreed with it, so option B is also incorrect. The other options, A and D, are not mutually exclusive, so let's analyze them further. Option A is telling us that Keynes refined classical economic theory, but this is only half the story. Keynes' work was actually a major departure from classical economics, as he rejected some of its fundamental assumptions. Option D is the most accurate description of Keynes' work. The Great Depression was a major economic crisis that occurred during Keynes' lifetime, and he wrote his most famous book, \textit{The General Theory of Employment, Interest and Money}, in part to explain the cause of the Depression and propose policies to address it. In fact, Keynesian economics is often associated with government intervention in the economy to stabilize output during recessions, which was a radical departure from classical economics.

\textbf{Answer:} D \\
\textbf{Correctly Applied Knowledge:} (2), (3), (4), (5), (6), (7), (8) \\
\textbf{KP:} 100\% \hspace{5em} \textbf{KR:} 75\% (6 out of 8) \\
\textbf{Final Answer: Correct}
\endgroup
\end{tcolorbox}
}
\captionsetup{skip=8pt, labelfont=bf}
\caption{Illustrative example of DPO training responses sampled by Knowledge Recall (KR). For the same question, a low-KR response and a high-KR response are compared.}
\label{table:minmax}
\end{table*} 

\clearpage
\begin{table*}[t]
    \centering
    \scalebox{1}{
        \begin{tcolorbox}[colback=gray!3,colframe=black]
        """ \\You are given a knowledge statement that is intended to help answer a specific question.\\
        Your task is to assess whether this statement is factually correct, based solely on objective and verifiable facts, using web search.\\\\
        * Focus only on the factual accuracy of the statement as written.\\
        * Do not evaluate how complete, useful, or relevant the statement is.\\
        * If there are no factually incorrect parts, consider the statement 'True'.\\
        * If there is any factual inaccuracy, mark it 'False'.\\\\
        Please follow these steps:\\
        1. Use web search to verify the statement.\\
        2. Write a detailed feedback explaining your fact-checking process and findings.\\
        3. Conclude with a final verdict: \texttt{True} if factually accurate, \texttt{False} otherwise.\\\\
        ---\\
        Output Format:\\
        
        \#\#\# Feedback:\\
        (write a fact-checking explanation based on search results) \\\\
        \#\#\# Evaluation:\\
        \texttt[True/False]\\\\
        ---\\
        Now, please evaluate the following statement based on the question provided:\\
        \#\#\# Corresponding Question:\\
        \{<question>\}\\\\
        \#\#\# Statement to Evaluate:\\
        \{<knowledge>\}\\
        """
        \end{tcolorbox}
    }
    \caption{Prompt used to evaluate the factual accuracy of PK units.}
\label{table:prompt_factuality}
\end{table*}
\begin{table*}[]
\centering
\resizebox{\textwidth}{!}{%
\begin{tabular}{lcccccc}
\toprule
\textbf{Category} & \textbf{True Negative} & \textbf{Omission Bias} & \textbf{Ambiguous Fact} & \textbf{Overly Strict Evaluation} & \textbf{Model Knowledge Deficiency} & \textbf{Total} \\
\midrule
\textbf{Sample Count} & 33 & 6 & 23 & 25 & 13 & 100 \\
\bottomrule
\end{tabular}%
}
\caption{Human evaluation of knowledge units initially judged as factually false. The table categorizes the types of errors identified across 100 sampled cases.}
\label{table:human_eval}
\end{table*}
\begin{table*}[t]
    \centering
    \scalebox{1}{
        \begin{tcolorbox}[colback=gray!3, colframe=black]
        """\\
        Evaluate the following knowledge strictly based on the following criterion: \textbf{Relevance}.\\
        \\
        1. Write a detailed feedback that assesses the quality of the required conceptual knowledge according to the given score rubric.\\
        2. After writing the feedback for the criterion, assign a score (an integer between 1 and 5) based on the rubric.\\
        3. The output format must be as follows:\\
        \texttt{Evaluation:}\\
        \texttt{Relevance: (detailed feedback) \texttt{Score: 1\string~5}}\\
        4. Please do not generate any other opening or closing remarks.\\\\
        ---\\
        \textbf{Input Format:}\\
        \#\#\# Question and Answer:\\
        Question: \{<question>\} \\
        Answer: \{<answer>\} \\\\
        \#\#\# Required Conceptual knowledge to evaluate:\\
        \{<knowledge>\} \\\\
        \#\#\# Score Rubric:\\
        \textbf{Relevance:} Does the knowledge represent essential background knowledge required to understand the question and reason about the answer?\\\\
        \textbf{Score 1:} The knowledge is completely irrelevant or unrelated to the question's context.\\
        \textbf{Score 2:} The knowledge is somewhat related to the question’s topic but provides very limited value in understanding the question.\\
        \textbf{Score 3:} The knowledge is generally related and helpful for understanding the question or reasoning about the answer.\\
        \textbf{Score 4:} The knowledge is closely related and highly valuable for clearly understanding the question’s context or the given answer choices.\\
        \textbf{Score 5:} The knowledge is essential, foundational background knowledge necessary for understanding the question and directly reasoning to the correct answer.\\
        ---\\
        \texttt{Evaluation:}\\
        """
        \end{tcolorbox}
    }
    \caption{Prompt used to evaluate the relevance of Principal Knowledge units.}
    \label{table:prompt_relevance}
\end{table*}


\clearpage
\begin{table*}[t]
    \centering
    \scalebox{1}{
        \begin{tcolorbox}[colback=gray!3,colframe=black]
        \textcolor{blue}{\textbf{[System Prompt]}}\\
        """You are an expert in identifying the conceptual knowledge required to answer a given question. \\
        Your task is to list all the essential pieces of conceptual knowledge needed to solve the question. \\
        Each piece of knowledge should be expressed as a single, clear, and concise sentence. \\
        
        Follow these steps: \\
        
        1. Identify the subject of the question. \\
        2. Break down the knowledge required to solve the question into concise, one-sentence explanations. \\
        3. Ensure that each sentence directly addresses a key concept, principle, or fact necessary for solving the question. \\
        4. Exclude knowledge that involves reasoning, process steps, or calculations specific to the problem. \\
        
        Your output should follow this format: \\
        
        \#\#\# Required Conceptual Knowledge \\
        1. "First concept as a single, concise sentence." \\
        2. "Second concept as a single, concise sentence." \\
        ...""" \\
        \\
        \textcolor{blue}{\textbf{[User Prompt]}}\\
        """\#\#\# Question \\
        \{<question>\} \\
        
        \#\#\# The Answer to the Question \\
        \{<answer>\} \\
        
        \#\#\# Required Conceptual Knowledge \\
        """ 
        \end{tcolorbox}
    }
    \caption{Prompt for extracting atomic knowledge elements from question-answer pairs. We guides the LLM to identify and enumerate relevant knowledge that aids in reasoning for the given question and answer. "<question>" and "<answer>" refers to the specific quesetion and answer, respectively.}
\label{table:prompt_knowledge}
\end{table*}
\begin{table*}[t]
    \centering
    \scalebox{1}{
        \begin{tcolorbox}[colback=gray!3,colframe=black]
        \textcolor{blue}{\textbf{[System Prompt]}}\\
        """
        You are an expert in analyzing rationales to evaluate the application of conceptual knowledge.\\
        Your task is to assess each piece of knowledge from the provided list and determine:\\
        1. Whether the knowledge was applied explicitly, implicitly, or not at all.\\
        2. Evaluate the correctness of the application.\\
        3. Explain the details in a single sentence.\\\\
        
        Provide your evaluation in the following format without :\\
        ---\\
        Concept: <concept text>\\
        Application: "explicit" or "implicit" or "unapplied"\\
        Correctness: "correct", "incorrect", or "N/A"\\
        Details:  "If explicit, specify the exact portion of the rationale. If implicit, explain how the rationale implies it. For unapplied, leave this blank or state 'Not mentioned in the rationale.'"\\
        ---\\
        """ \\
        \\
        \textcolor{blue}{\textbf{[User Prompt]}}\\
        """
        \#\#\# Question\\
        \{<question>\}\\
        \\
        \#\#\# The Answer to the Question\\
        \{<answer>\}\\
        \\
        \#\#\# Required Conceptual Knowledge\\
        \{<knowledge>\}\\
        \\
        \#\#\# Rationale\\
        \{<rationale>\}\\
        \\
        \#\#\# Evaluation\\
        """
        \end{tcolorbox}
    }
    \caption{Prompt for evaluating the application of atomic knowledge. We assess whether each piece of relevant knowledge is applied explicitly, implicitly, or not at all, and to evaluate its correctness based on the provided rationale.}
\label{table:prompt_evaluation}
\end{table*}
\clearpage
\begin{table*}[]
\small
\centering
\begin{tabular}{lccccc} 
\toprule
 & KP & KR & F1 & Acc. & Token length \\ 
\midrule
zero-shot CoT & 92.3 & 82.6 & 86.9 & 66.3 & 238.4 \\
Answer-driven & 94.4 {\scriptsize (\footnotesize +2.1\%)} & 85.9 {\scriptsize (\footnotesize +3.3\%)} & 88.6 {\scriptsize (\footnotesize +1.7\%)} & 70.3 {\scriptsize (\footnotesize +4.0\%)} & 271.1 {\scriptsize (\footnotesize +13.7\%)} \\

Random KR & 96.2 {\scriptsize (\footnotesize +3.9\%)} & 87.6 {\scriptsize (\footnotesize +5.0\%)} & 89.5 {\scriptsize (\footnotesize +2.7\%)} & 70.8 {\scriptsize (\footnotesize +4.5\%)} & 313.4 {\scriptsize (\footnotesize +31.5\%)} \\

Max KR & 96.5 {\scriptsize (\footnotesize +4.2\%)} & 92.6 {\scriptsize (\footnotesize +10.0\%)} & 94.1 {\scriptsize (\footnotesize +7.2\%)} & 70.1 {\scriptsize (\footnotesize +3.8\%)} & 378.0 {\scriptsize (\footnotesize +58.6\%)} \\
Min KR & 94.8 {\scriptsize (\footnotesize +2.5\%)} & 74.4 {\scriptsize (\footnotesize -8.2\%)} & 77.2 {\scriptsize (\footnotesize -9.7\%)} & 70.6 {\scriptsize (\footnotesize +4.4\%)} & 198.6 {\scriptsize (\footnotesize -16.7\%)} \\
Max length & 95.7 {\scriptsize (\footnotesize +3.3\%)} & 91.6 {\scriptsize (\footnotesize +9.0\%)} & 93.6 {\scriptsize (\footnotesize +6.7\%)} & 67.3 {\scriptsize (\footnotesize +1.0\%)} & 428.7 {\scriptsize (\footnotesize +79.9\%)} \\
Min length & 78.2 {\scriptsize (\footnotesize -14.1\%)} & 52.4 {\scriptsize (\footnotesize -30.2\%)} & 60.1 {\scriptsize (\footnotesize -26.8\%)} & 62.7 {\scriptsize (\footnotesize -3.6\%)} & 38.5 {\scriptsize (\footnotesize -83.8\%)}\\ 
\bottomrule
\end{tabular}
\caption{Preference Optimization results including length based data selection}
\label{length_based_performance}
\end{table*}

\begin{table*}[]
\small
\centering
\resizebox{\textwidth}{!}{%
\begin{tabular}{clcl} \toprule
\textbf{Category} & \textbf{Subject} & \textbf{Category} & \textbf{Subject} \\ \midrule
STEM & abstract\_algebra & Humanities & formal\_logic \\
 & anatomy & & high\_school\_european\_history \\
 & astronomy & &high\_school\_us\_history \\
 & college\_biology & &high\_school\_world\_history \\
 & college\_chemistry & &international\_law \\
 & college\_computer\_science & &jurisprudence \\
 & college\_mathematics & &logical\_fallacies \\
 & college\_physics & & moral\_disputes \\
 & computer\_security & &moral\_scenarios \\
 & conceptual\_physics & &philosophy \\
 & electrical\_engineering & &prehistory \\
 & elementary\_mathematics & &professional\_law \\
 & high\_school\_biology & & world\_religions \\
 & high\_school\_chemistry \\
 & high\_school\_computer\_science \\
 & high\_school\_mathematics \\
 & high\_school\_physics \\
 & high\_school\_statistics \\
 & machine\_learning \\ \midrule

Other & business\_ethics & Social Sciences & econometrics \\
 & clinical\_knowledge &  & high\_school\_geography \\
 & college\_medicine & & high\_school\_government\_and\_politics \\
 & global\_facts & & high\_school\_macroeconomics \\
 & human\_aging & & high\_school\_microeconomics \\
 & management & & high\_school\_psychology \\
 & marketing & & human\_sexuality \\
 & medical\_genetics & & professional\_psychology \\
 & miscellaneous &  & public\_relations \\
 & nutrition &  & security\_studies \\
 & professional\_accounting &  & sociology \\
 & professional\_medicine &  & sociology \\
 & virology & & \\ \bottomrule
\end{tabular}%
}
\caption{Category classification of MMLU subjects}
\label{tab:subjects}
\end{table*}
\begin{table*}
\small
\centering
{%
\begin{tabular}{lr} \toprule
 & \textbf{Sample count / Ratio} \\ \midrule
DeepSeek-R1-Distill-Llama-8B & 844 / 6.01\% \\
DeepSeek-R1-Distill-Llama-70B & 194 / 1.38\% \\
DeepSeek-R1-Distill-Qwen-7B & 1048 / 7.46\% \\
DeepSeek-R1-Distill-Qwen-32B & 352 / 2.51\% \\ \bottomrule
\end{tabular}%
}
\caption{Counts and ratio of omitted sample due to token length}
\label{deepseek_statistics}
\end{table*}

\begin{table*}
\centering
\begin{minipage}{0.45\textwidth}
\resizebox{\columnwidth}{!}{%
\begin{tabular}{lccccc} 
\toprule
\textbf{\;STEM\qquad\quad} & \textbf{KP} & \textbf{KR} & \textbf{F1} & \textbf{Acc.} & \textbf{Token length} \\ \midrule
zero-shot CoT & 91.8 & 80.9 & 84.2 & 60.6 & 264.8 \\
Answer-driven & 94.3 & 83.7 & 87.2 & 63.1 & 287.3 \\
Random KR & 95.9 & 86.3 & 89.6 & 63.5 & 363.9 \\
Max KR & 96.3 & 90.9 & 92.7 & 64.6 & 403.5 \\
Min KR & 94.6 & 76.0 & 82.1 & 66.1 & 226.7 \\
Max length & 95.3 & 89.1 & 91.1 & 59.7 & 484.0 \\
Min length & 70.9 & 46.8 & 53.9 & 54.2 & 31.7\\ \bottomrule
\end{tabular}%
}
\label{tab:llama_dpo_1}
\end{minipage}%
\hfill
\begin{minipage}{0.45\textwidth}
\resizebox{\columnwidth}{!}{%
\begin{tabular}{lccccc} \toprule
\textbf{Humanities} & \textbf{KP} & \textbf{KR} & \textbf{F1} & \textbf{Acc.} & \textbf{Token length} \\ \midrule
zero-shot CoT & 90.3 & 84.4 & 85.6 & 59.9 & 248.3 \\
Answer-driven & 64.2 & 93.4 & 87.6 & 89.1 & 293.1 \\
Random KR & 95.9 & 89.6 & 91.6 & 66.2 & 326.8 \\
Max KR & 95.8 & 93.4 & 93.9 & 65.2 & 393.8 \\
Min KR & 93.9 & 78.7 & 83.6 & 65.7 & 221.1 \\
Max length & 95.2 & 92.6 & 93.0 & 61.0 & 442.0 \\
Min length & 56.9 & 80.5 & 60.8 & 67.1 & 39.6\\ \bottomrule
\end{tabular}%
}
\label{tab:humanities_llama_dpo_1}
\end{minipage}
\vspace{1em}

\begin{minipage}{0.45\textwidth}
\centering
\resizebox{\columnwidth}{!}{%
\begin{tabular}{lccccc} \toprule
\textbf{Social Sciences} & \textbf{KP} & \textbf{KR} & \textbf{F1} & \textbf{Acc.} & \textbf{Token length} \\ \midrule
zero-shot CoT & 94.4 & 84.0 & 87.1 & 74.0 & 228.9 \\
Answer-driven & 77.4 & 95.0 & 86.8 & 89.4 & 253.4 \\
Random KR & 96.1 & 87.6 & 90.5 & 77.4 & 295.8 \\
Max KR & 97.5 & 93.3 & 94.7 & 77.1 & 374.7 \\
Min KR & 95.2 & 72.2 & 79.8 & 76.6 & 176.9 \\
Max length & 96.2 & 92.9 & 93.9 & 75.6 & 401.8 \\
Min length & 72.4 & 84.5 & 53.7 & 62.6 & 50.0\\ \bottomrule
\end{tabular}%
}
\label{tab:llama_dpo_2}
\end{minipage}%
\hfill
\begin{minipage}{0.45\textwidth}
\resizebox{\columnwidth}{!}{%
\begin{tabular}{lccccc} \toprule
\textbf{\;Others\qquad} & \textbf{KP} & \textbf{KR} & \textbf{F1} & \textbf{Acc.} & \textbf{Token length} \\ \midrule
zero-shot CoT & 94.1 & 80.3 & 84.6 & 74.0 & 206.0 \\
Answer-driven & 95.7 & 84.5 & 88.3 & 79.6 & 239.0 \\
Random KR & 97.1 & 85.8 & 89.9 & 77.6 & 260.9 \\
Max KR & 96.8 & 92.4 & 94.0 & 76.2 & 331.5 \\
Min KR & 96.0 & 68.4 & 77.3 & 76.8 & 157.4 \\
Max length & 96.3 & 91.5 & 93.1 & 76.2 & 379.2 \\
Min length & 76.0 & 44.1 & 53.1 & 70.4 & 32.5\\ \bottomrule
\end{tabular}%
}
\label{tab:llama_dpo_3}
\end{minipage}
\caption{Categorywise Preference Optimization results of Llama3-8B-Instruct}
\end{table*}

\end{document}